\documentclass[conference]{IEEEtran}
\IEEEoverridecommandlockouts
\usepackage{orcidlink}
\usepackage{cite}
\usepackage{amsmath,amssymb,amsfonts}
\usepackage{graphicx}
\usepackage{textcomp}
\usepackage{xcolor}
\usepackage[caption=false]{subfig}
\usepackage{booktabs}
\usepackage{multirow}
\usepackage{url}
\usepackage{listings}

\usepackage{tikz}
\usetikzlibrary{shapes.geometric, arrows, positioning, fit, backgrounds, calc}

\definecolor{apiblue}{HTML}{E1F5FE}
\definecolor{apiborder}{HTML}{01579B}
\definecolor{coreorange}{HTML}{FFF3E0}
\definecolor{coreborder}{HTML}{E65100}
\definecolor{datapurple}{HTML}{F3E5F5}
\definecolor{databorder}{HTML}{7B1FA2}

\def\BibTeX{{\rm B\kern-.05em{\sc i\kern-.025em b}\kern-.08em
    T\kern-.1667em\lower.7ex\hbox{E}\kern-.125emX}}
\begin{document}

\title{ART:  Adaptive Response Tuning Framework\\
\large A Multi-Agent Tournament-Based Approach to LLM Response Optimization}

\author{%
    \IEEEauthorblockN{Omer Jauhar Khan \orcidlink{0009-0006-4203-4665}}
    \IEEEauthorblockA{\textit{Department of Computer Science}\\
    \textit{National University of Computer and Emerging Sciences (FAST-NUCES)}\\
    Peshawar, 25000, Pakistan\\
    Email: p218055@pwr.nu. edu.pk}
}

\maketitle

\begin{abstract}
Large Language Models (LLMs) have demonstrated remarkable capabilities in natural language understanding and generation. However, single-model responses often exhibit inconsistencies, hallucinations, and varying quality across different query domains. This paper presents \textbf{ART (Adaptive Response Tuning)}, a novel framework that employs tournament-style ELO ranking and multi-agent reasoning to systematically optimize LLM outputs. By enabling multiple LLM agents to compete, critique, and collaborate through structured tournament workflows, ART produces consensus responses that outperform individual model outputs. Our framework introduces configurable tournament parameters, dynamic agent selection, and multiple consensus fusion strategies. Experimental evaluations demonstrate significant improvements in response accuracy, coherence, and reliability compared to baseline single-model approaches. The ART framework provides a scalable, production-ready solution for applications requiring high-quality, vetted LLM responses, achieving an 8.4\% improvement in overall quality metrics and R² values exceeding 0.96 in ELO rating convergence.
\end{abstract}

\begin{IEEEkeywords}
Large Language Models, Multi-Agent Systems, ELO Rating, Tournament Selection, Consensus Generation, Response Optimization, Natural Language Processing, LLM Evaluation
\end{IEEEkeywords}
\section{Introduction}

\subsection{Background and Motivation}

The proliferation of Large Language Models (LLMs) has revolutionized natural language processing, enabling sophisticated applications ranging from conversational agents to code generation and scientific analysis. Models such as GPT-4, Claude, LLaMA, and PaLM demonstrate impressive capabilities in understanding context, generating coherent text, and reasoning about complex topics~\cite{brown2020language,ouyang2022training}.

However, despite these advances, individual LLM responses suffer from several well-documented limitations:

\begin{enumerate}
    \item \textbf{Hallucinations}: LLMs may generate plausible-sounding but factually incorrect information~\cite{ji2023survey}
    \item \textbf{Inconsistency}: Repeated queries may yield different responses of varying quality~\cite{elazar2021measuring}
    \item \textbf{Bias}: Individual models may exhibit systematic biases from their training data~\cite{bender2021dangers}
    \item \textbf{Domain Limitations}: Models may excel in certain domains while underperforming in others~\cite{wang2022self}
    \item \textbf{Confidence Calibration}: LLMs often struggle to accurately assess their own uncertainty~\cite{xiong2023can}
\end{enumerate}

These limitations pose significant challenges for applications requiring reliable, accurate, and consistent outputs. Medical diagnosis assistance, legal document analysis, educational content generation, and financial decision support are examples where response quality is critical.

\subsection{Problem Statement}

Given a query $Q$ and a set of $n$ LLM agents $A = \{a_1, a_2, ..., a_n\}$, each capable of generating a response $r_i = a_i(Q)$, the problem is to:

\begin{enumerate}
    \item \textbf{Evaluate} the quality of each response $r_i$ across multiple criteria
    \item \textbf{Rank} agents based on their response quality using a principled scoring system
    \item \textbf{Select or synthesize} an optimal response $R^*$ that maximizes overall quality
    \item \textbf{Adapt} agent rankings over time to reflect cumulative performance
\end{enumerate}

\subsection{Contributions}

This paper makes the following contributions:

\begin{enumerate}
    \item \textbf{ART Framework Architecture}: A comprehensive multi-agent framework for LLM response optimization featuring modular components for agent management, tournament orchestration, and consensus generation.
    
    \item \textbf{Tournament-Based ELO Ranking}: Application of the ELO rating system to LLM agent evaluation, with extensions for multi-agent matches, partial wins, and dynamic K-factor adjustment.
    
    \item \textbf{Multi-Strategy Consensus Engine}: Multiple fusion strategies for synthesizing optimal responses from agent outputs, including weighted voting, contextual aggregation, and hybrid synthesis.
    
    \item \textbf{Empirical Evaluation}: Comprehensive experiments demonstrating the effectiveness of tournament-based optimization across diverse query types and model configurations.
    
    \item \textbf{Production-Ready Implementation}: A complete, documented implementation with RESTful API, Docker deployment, and extensive test coverage.
\end{enumerate}

\subsection{Paper Organization}

The remainder of this paper is organized as follows: Section~\ref{sec:related} reviews related work in multi-agent LLM systems and response optimization. Section~\ref{sec:methodology} presents the theoretical foundations of the ART framework. Section~\ref{sec:architecture} details the system architecture and implementation. Section~\ref{sec:experiments} describes the experimental methodology and presents results. Section~\ref{sec:discussion} discusses implications and limitations. Section~\ref{sec:conclusion} concludes with future research directions.
\section{Related Work}
\label{sec:related}

\subsection{Multi-Agent LLM Systems}

The use of multiple LLM agents for improved task performance has gained significant attention. \textbf{Debate frameworks} enable agents to argue different positions, leading to more nuanced responses~\cite{du2023improving}. \textbf{Self-consistency} approaches generate multiple responses and select based on majority voting~\cite{wang2022self}. \textbf{Chain-of-thought} prompting encourages step-by-step reasoning that can be verified by other agents~\cite{wei2022chain}.

Notable works include:
\begin{itemize}
    \item \textbf{Constitutional AI} (Anthropic): Uses principles for self-critique~\cite{bai2022constitutional}
    \item \textbf{Tree of Thoughts}: Explores multiple reasoning paths~\cite{yao2023tree}
    \item \textbf{LLM Cascades}: Routes queries to appropriate models~\cite{chen2023frugalgpt}
    \item \textbf{Mixture of Experts}: Combines specialized model outputs~\cite{shazeer2017outrageously}
\end{itemize}

ART differs from these approaches by introducing explicit competitive evaluation through tournament mechanics and persistent ELO-based ranking.

\subsection{Response Quality Evaluation}

Automated evaluation of LLM responses remains challenging. Existing approaches include:

\begin{itemize}
    \item \textbf{Reference-based metrics}: BLEU, ROUGE, BERTScore compare against gold standards~\cite{papineni2002bleu,lin2004rouge}
    \item \textbf{Reference-free metrics}: Perplexity, coherence scores, factual consistency checks~\cite{guan2021keywords}
    \item \textbf{Human evaluation}: Gold standard but expensive and time-consuming~\cite{clark2021all}
    \item \textbf{LLM-as-judge}: Using another LLM to evaluate responses~\cite{zheng2023judging}
\end{itemize}

ART employs LLM-based cross-evaluation where agents critique each other's responses, providing multi-perspective assessment without requiring external references.

\subsection{ELO Rating Systems}

The ELO rating system, developed by Arpad Elo for chess rankings~\cite{elo1978rating}, provides a principled approach to relative skill estimation. The system has been applied to various domains including video games, sports, and academic competitions.

In AI systems, ELO has been used for:
\begin{itemize}
    \item AlphaGo agent evaluation~\cite{silver2016mastering}
    \item Chatbot Arena LLM rankings~\cite{zheng2023judging}
    \item Reinforcement learning agent comparison~\cite{czarnecki2020real}
\end{itemize}

ART extends traditional ELO with support for multi-agent matches and partial win scores based on response quality differentials.

\subsection{Consensus and Voting Systems}

Aggregating multiple opinions to reach consensus is well-studied in distributed systems and social choice theory~\cite{arrow2012social}:

\begin{itemize}
    \item \textbf{Majority voting}: Simple but ignores expertise differences
    \item \textbf{Weighted voting}: Accounts for agent reliability~\cite{raykar2010learning}
    \item \textbf{Bayesian aggregation}: Probabilistic combination~\cite{yi2012inferring}
    \item \textbf{Deliberation}: Iterative refinement through discussion~\cite{liang2023encouraging}
\end{itemize}

ART implements multiple consensus strategies, allowing selection based on application requirements.

\section{Theoretical Foundations}
\label{sec:methodology}

\subsection{ELO Rating System}

The ELO rating system estimates the relative skill of players based on match outcomes. For two players with ratings $R_A$ and $R_B$, the expected score for player A is:

\begin{equation}
E_A = \frac{1}{1 + 10^{(R_B - R_A)/400}}
\end{equation}

After a match, ratings are updated:

\begin{equation}
R'_A = R_A + K \cdot (S_A - E_A)
\end{equation}

Where:
\begin{itemize}
    \item $K$ is the K-factor determining update sensitivity
    \item $S_A$ is the actual score (1 for win, 0.5 for draw, 0 for loss)
    \item $E_A$ is the expected score
\end{itemize}

\begin{figure*}[!htbp]
\centering
\includegraphics[width=\textwidth]{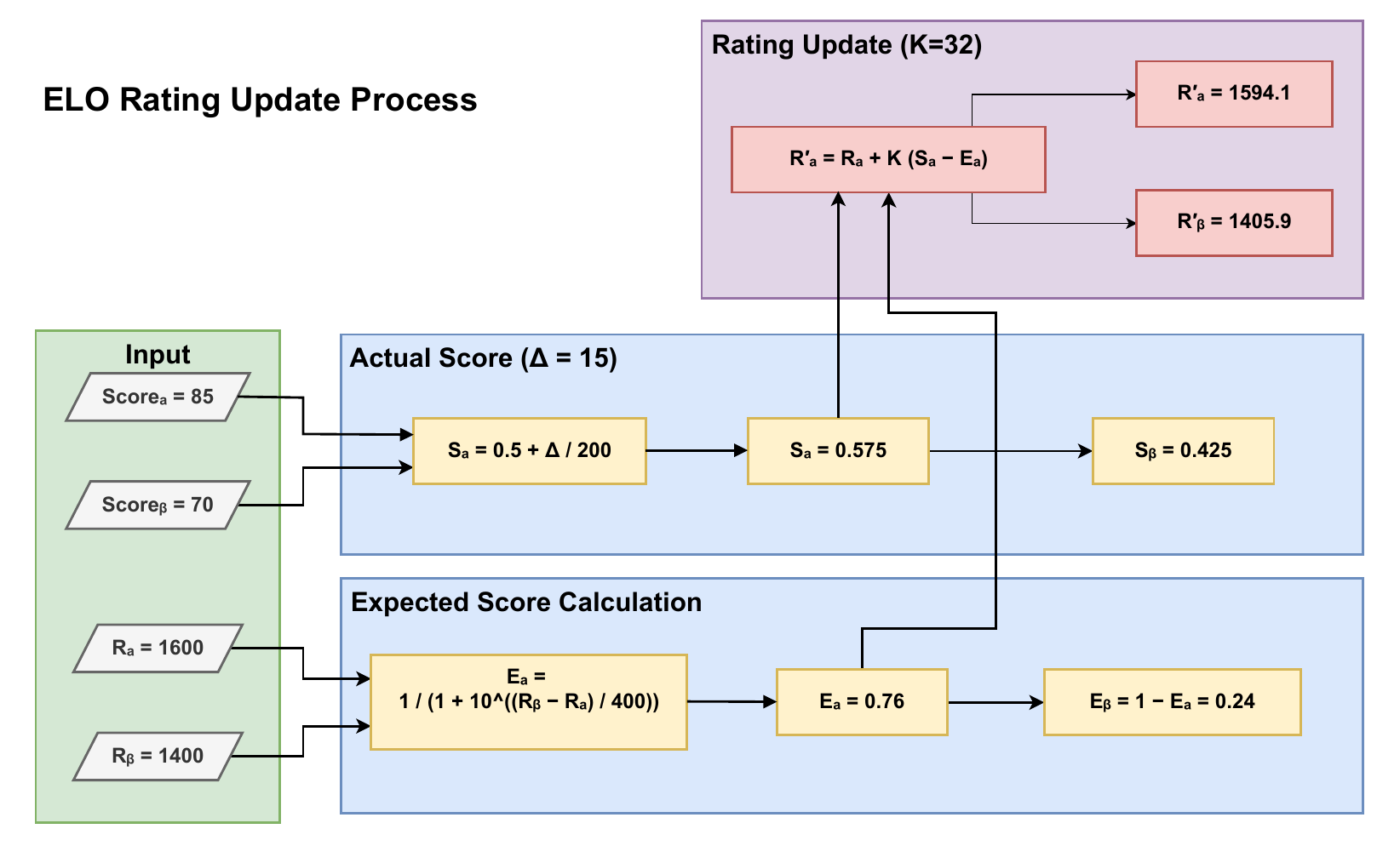}
\caption{ELO rating update process flowchart demonstrating score calculation. }
\label{fig:elo_process}
\end{figure*}

% IMPROVEMENT: Add text reference to the figure
The ELO rating update process is illustrated in Fig.~\ref{fig:elo_process}, showing how expected scores are computed and ratings adjusted based on match outcomes.

\subsection{Extensions for Multi-Agent Response Evaluation}

\subsubsection{Weighted Scoring}

Traditional ELO uses binary outcomes. ART extends this with continuous scoring based on response quality differentials: 

\begin{equation}
S_A = \begin{cases}
0.5 + \frac{\Delta_{AB}}{200} & \text{if } |\Delta_{AB}| > \theta \\
0.5 & \text{otherwise (draw)}
\end{cases}
\end{equation}

Where $\Delta_{AB} = Q_A - Q_B$ is the quality score differential and $\theta$ is the draw threshold.

\subsubsection{Multi-Agent Matches}

For $n > 2$ agents, ART employs round-robin pairwise comparison with adjusted K-factor:

\begin{equation}
K_{adj} = \frac{K}{n-1}
\end{equation}

\begin{figure*}[!htbp]
\centering
\includegraphics[width=\textwidth]{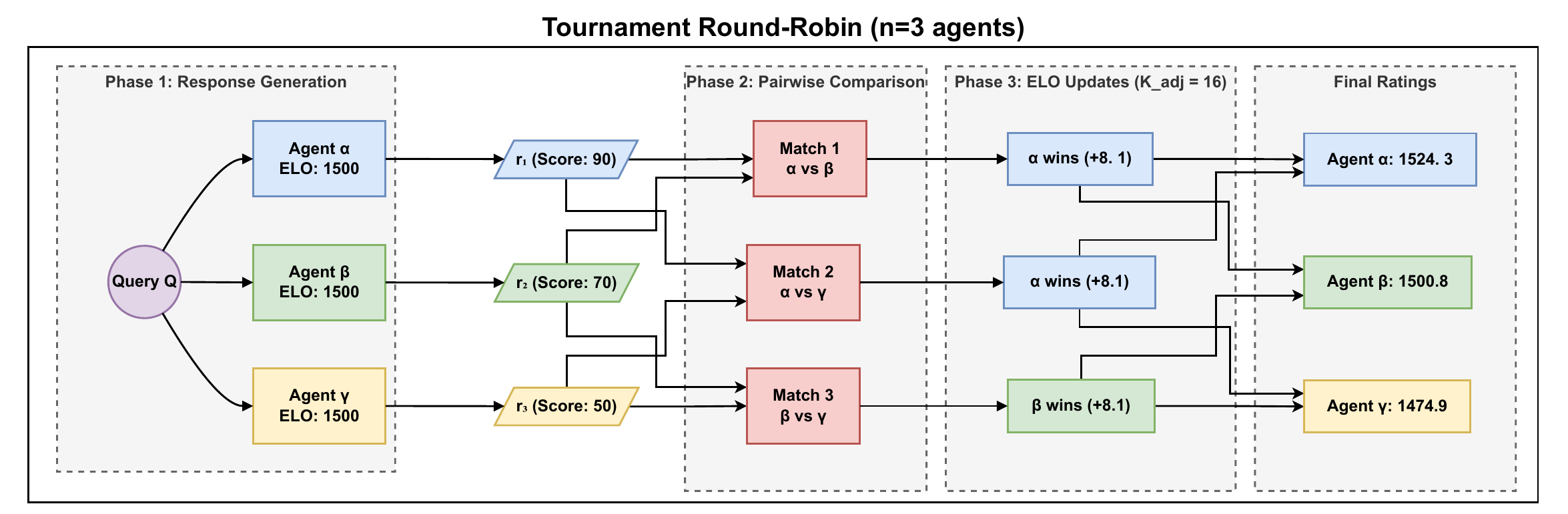}
\caption{Multi-agent round-robin tournament structure showing pairwise comparisons and ELO updates for three agents.}
\label{fig:roundrobin}
\end{figure*}

This prevents excessive rating volatility when many agents compete simultaneously.  An example tournament with three agents is shown in Fig.~\ref{fig:roundrobin}, demonstrating how pairwise matches yield cumulative ELO changes with the adjusted K-factor.

\subsubsection{Dynamic K-Factor}

The K-factor can be adjusted based on agent experience:

\begin{equation}
K_i = K_{base} \cdot \max(0.5, 1 - 0.1 \cdot \log(m_i + 1))
\end{equation}

Where $m_i$ is agent $i$'s match count. This reduces rating volatility for experienced agents. 

\subsection{Response Quality Metrics}

ART evaluates responses across four primary dimensions:

\begin{enumerate}
    \item \textbf{Accuracy} ($\alpha$): Factual correctness and reliability
    \item \textbf{Coherence} ($\gamma$): Logical flow and clarity
    \item \textbf{Completeness} ($\kappa$): Coverage of relevant aspects
    \item \textbf{Relevance} ($\rho$): Alignment with query intent
\end{enumerate}

The composite quality score is:

\begin{equation}
\label{eq:quality_score}
Q = w_\alpha \cdot \alpha + w_\gamma \cdot \gamma + w_\kappa \cdot \kappa + w_\rho \cdot \rho
\end{equation}

Where weights $w_x$ are configurable and sum to 1. 

\begin{figure}[!t]
\centering
\includegraphics[width=0.7\columnwidth]{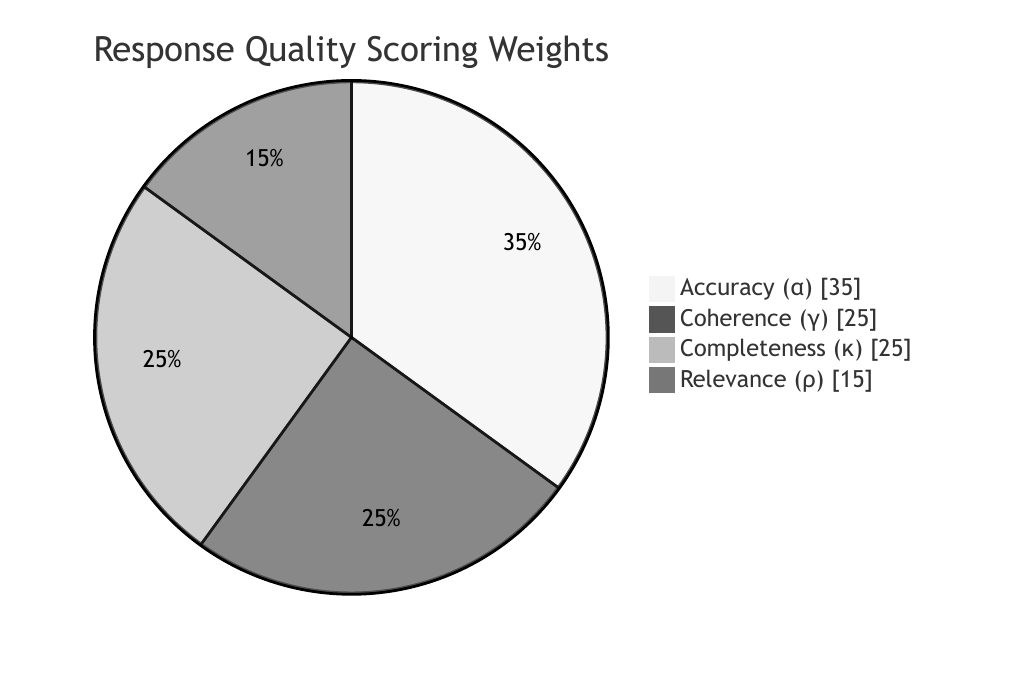}
\caption{Response quality metric weight distribution showing the relative importance of each evaluation dimension.}
\label{fig:quality_weights}
\end{figure}

% IMPROVEMENT: Add Diagram 6 here
The default weight distribution is visualized in Fig.~\ref{fig:quality_weights}, with accuracy receiving the highest weight (0.35) due to its critical importance in response evaluation.

\subsection{Consensus Generation}

ART implements multiple consensus strategies for synthesizing optimal responses from agent outputs.  The choice of strategy depends on application requirements and agent diversity.

\subsubsection{Weighted Voting}

Select response with maximum weighted support:

\begin{equation}
R^* = \arg\max_{r_i} \sum_{j \neq i} w_j \cdot \text{support}(r_j, r_i)
\end{equation}

Where $w_j$ is derived from ELO scores:

\begin{equation}
w_j = \frac{\text{normalize}(R_j)}{\sum_k \text{normalize}(R_k)}
\end{equation}

\subsubsection{Contextual Aggregation}

Combine complementary aspects from top-$k$ responses:

\begin{equation}
R^* = \text{Aggregate}(r_1, r_2, ..., r_k; w_1, w_2, ..., w_k)
\end{equation}

This preserves unique insights from multiple perspectives.

\subsubsection{Hybrid Synthesis}

Generate a new response using top responses as context: 

\begin{equation}
R^* = \text{LLM}(\text{Synthesize} | r_1, r_2, ..., r_k)
\end{equation}

This can produce responses superior to any individual input. 
\section{System Architecture}
\label{sec:architecture}

\subsection{Overview}

\begin{figure*}[!htbp]
\centering
\includegraphics[width=0.75\textwidth]{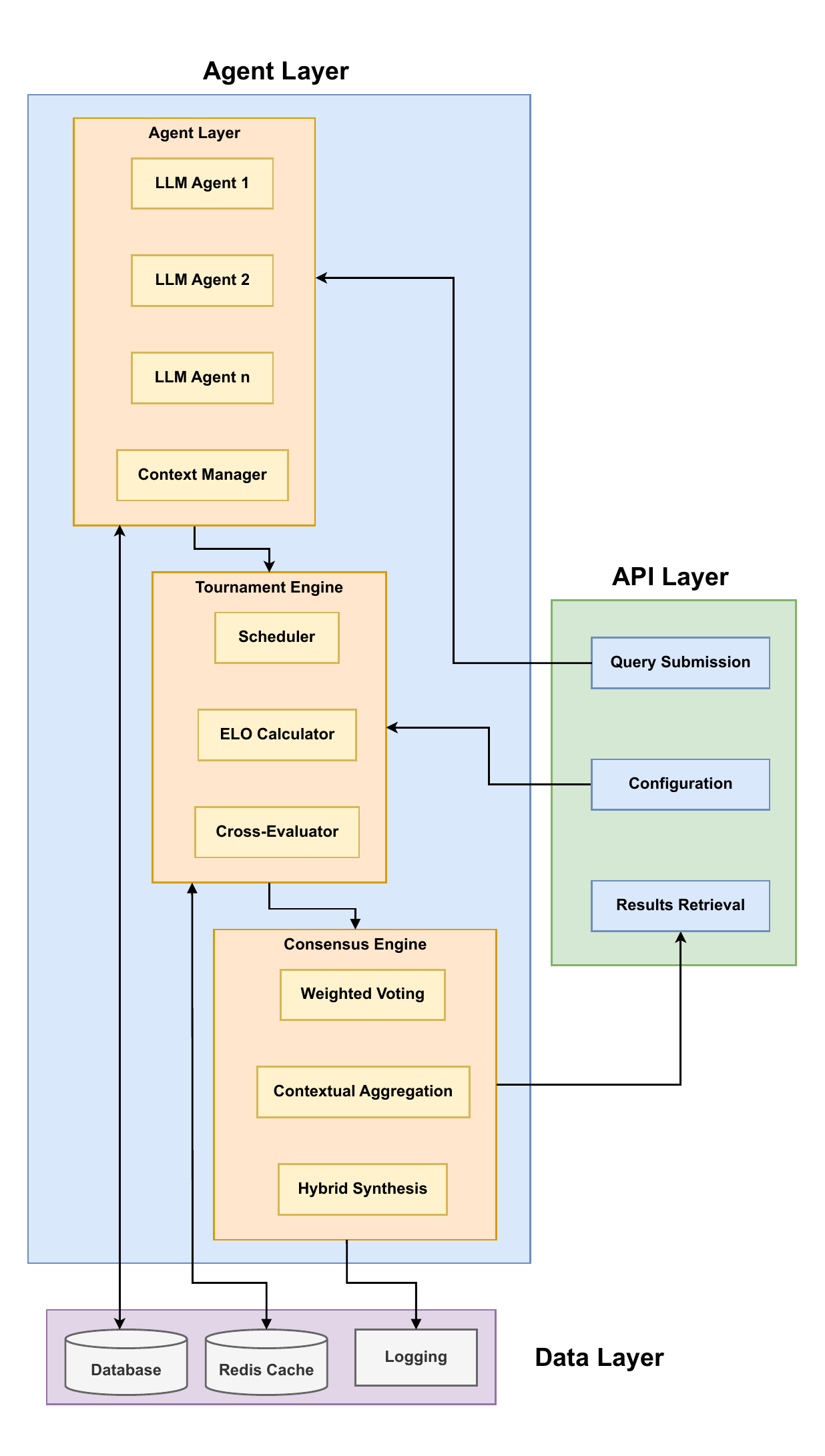}
\caption{ART framework architecture showing the layered organization of components.  The API Layer handles external interactions, the Core Processing Layer contains Agent, Tournament, and Consensus subsystems, and the Data Layer manages persistence, caching, and logging.}
\label{fig:architecture}
\end{figure*}

The ART framework consists of five primary components organized in a layered architecture as shown in Fig. ~\ref{fig:architecture}. 

\subsection{Agent Layer}

\subsubsection{Agent Interface}

Each agent implements a standardized interface:

\begin{itemize}
    \item \texttt{generate\_response(query, context)}: Produces initial response
    \item \texttt{critique\_response(query, response)}: Evaluates another agent's response
    \item \texttt{improve\_response(response, critiques)}: Refines based on feedback
\end{itemize}

\subsubsection{Agent State}

Agents maintain persistent state including: 
\begin{itemize}
    \item ELO score (initialized at 1500.0)
    \item Match history (wins, losses, draws)
    \item Context history for learning
    \item Performance metrics across query types
\end{itemize}

\begin{figure*}[!htbp]
\centering
\includegraphics[width=\textwidth]{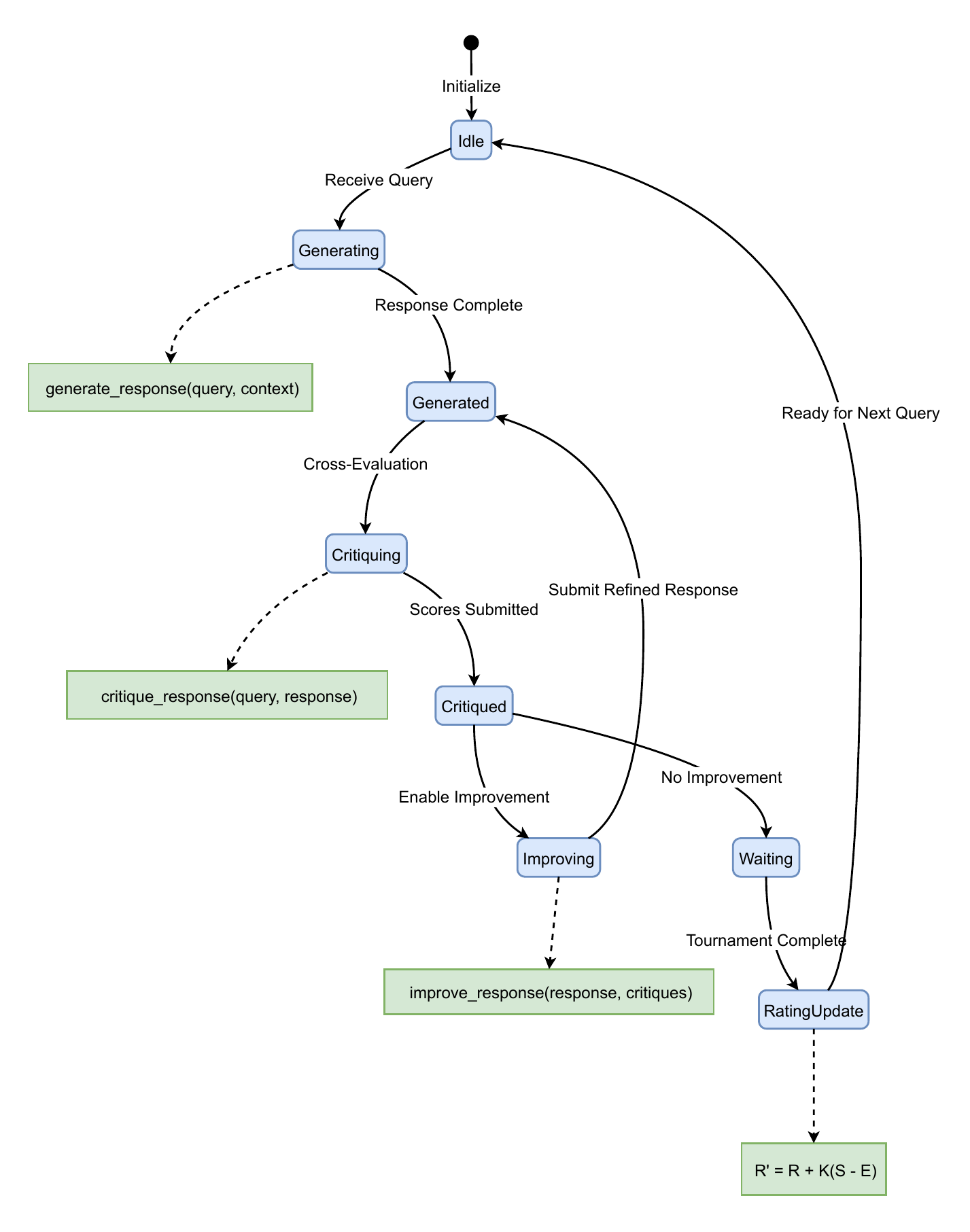}
\caption{Agent state machine depicting lifecycle transitions during tournament execution, including idle, generating, critiquing, improving, and rating update states.}
\label{fig:agent_state}
\end{figure*}

The agent lifecycle and state transitions during tournament execution are depicted in Fig.~\ref{fig:agent_state}, showing how agents progress from idle state through response generation, cross-evaluation, optional improvement, and rating updates.

\subsection{Tournament Engine}

\subsubsection{Tournament Configuration}

Configurable parameters include:
\begin{itemize}
    \item Number of tournaments per query
    \item Rounds per tournament
    \item K-factor for ELO updates
    \item Scoring weights for quality dimensions
    \item Response improvement enablement
    \item Dynamic agent selection
\end{itemize}

\subsubsection{Tournament Workflow}

\begin{figure*}[!t]
\centering
\includegraphics[width=\textwidth]{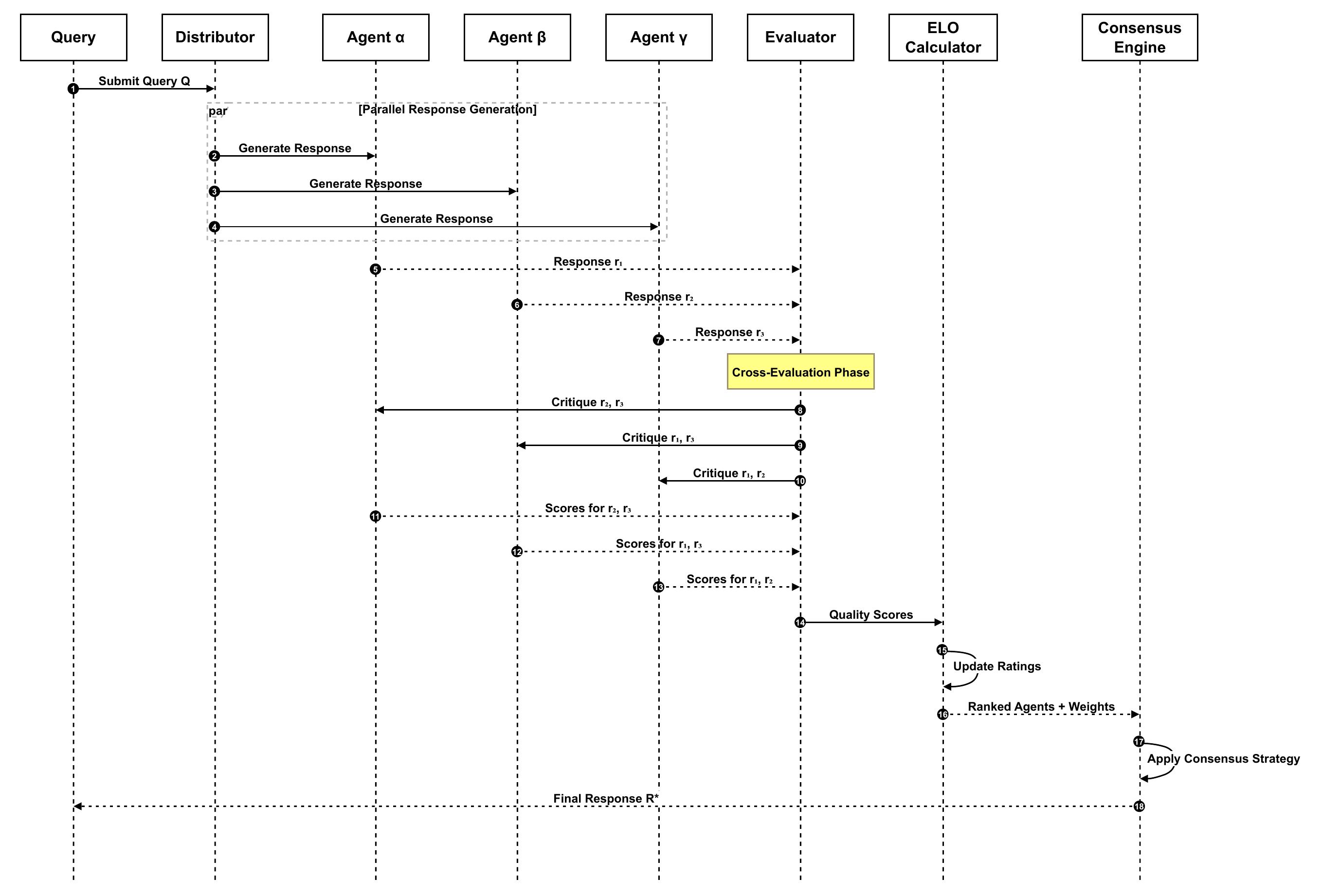}
\caption{Tournament workflow sequence diagram illustrating the seven-phase evaluation process from query distribution through consensus generation.}
\label{fig:tournament_workflow}
\end{figure*}

The tournament proceeds through structured phases as illustrated in Fig.~\ref{fig:tournament_workflow}:

\begin{enumerate}
    \item \textbf{Query Distribution}: Distribute query to all active agents
    \item \textbf{Response Generation}: Parallel response generation
    \item \textbf{Cross-Evaluation}:  Agents critique all other responses
    \item \textbf{Scoring}: Calculate weighted quality scores
    \item \textbf{ELO Update}: Update agent ratings
    \item \textbf{Response Improvement}: Optional iterative refinement
    \item \textbf{Consensus}: Generate final response
\end{enumerate}

\subsection{ELO Calculator}

The ELO calculator implements:
\begin{itemize}
    \item Expected score computation
    \item Rating update calculation
    \item Pairwise match processing
    \item Multi-agent match handling with adjusted K-factors
\end{itemize}

\subsection{Consensus Engine}

The consensus engine supports multiple strategies: 

\begin{itemize}
    \item \textbf{Top Response Selection}: Choose highest-rated response
    \item \textbf{Weighted Voting Fusion}: Combine based on ELO weights
    \item \textbf{Contextual Aggregation}: Merge complementary aspects
    \item \textbf{Hybrid Synthesis}: Generate new response from top-k
\end{itemize}

\begin{figure*}[! t]
\centering
\includegraphics[width=\textwidth]{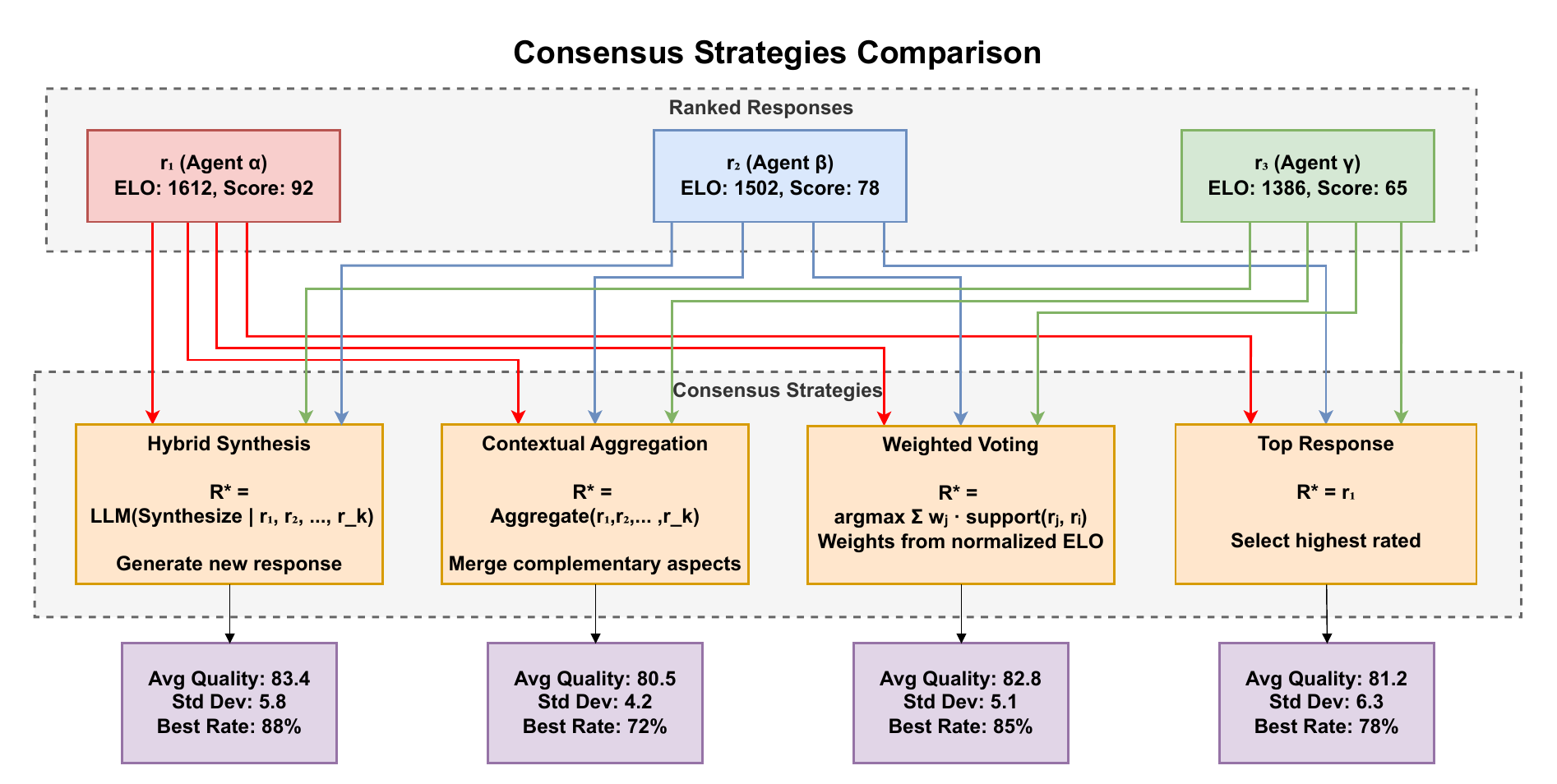}
\caption{Comparison of consensus generation strategies showing the four approaches (Top Response, Weighted Voting, Contextual Aggregation, and Hybrid Synthesis) with their respective quality metrics and selection criteria.}
\label{fig:consensus_strategies}
\end{figure*}

Strategy selection adapts to query characteristics and agent diversity.  A comparison of these consensus strategies with their performance characteristics is presented in Fig.~\ref{fig:consensus_strategies}.

\subsection{API Layer}

The framework exposes a RESTful API with endpoints for:

\begin{itemize}
    \item Query submission and retrieval
    \item Tournament execution and monitoring
    \item Agent registration and management
    \item Leaderboard access
    \item Configuration management
    \item System health and metrics
\end{itemize}

\subsection{Implementation Details}

The ART framework is implemented in Python with:
\begin{itemize}
    \item Asynchronous processing for parallel agent operations
    \item Database persistence for agent state and history
    \item Redis caching for performance optimization
    \item Comprehensive logging and monitoring
    \item Docker containerization for deployment
    \item Extensive unit and integration test coverage
\end{itemize}

\begin{figure*}[!t]
\centering
\includegraphics[width=\textwidth]{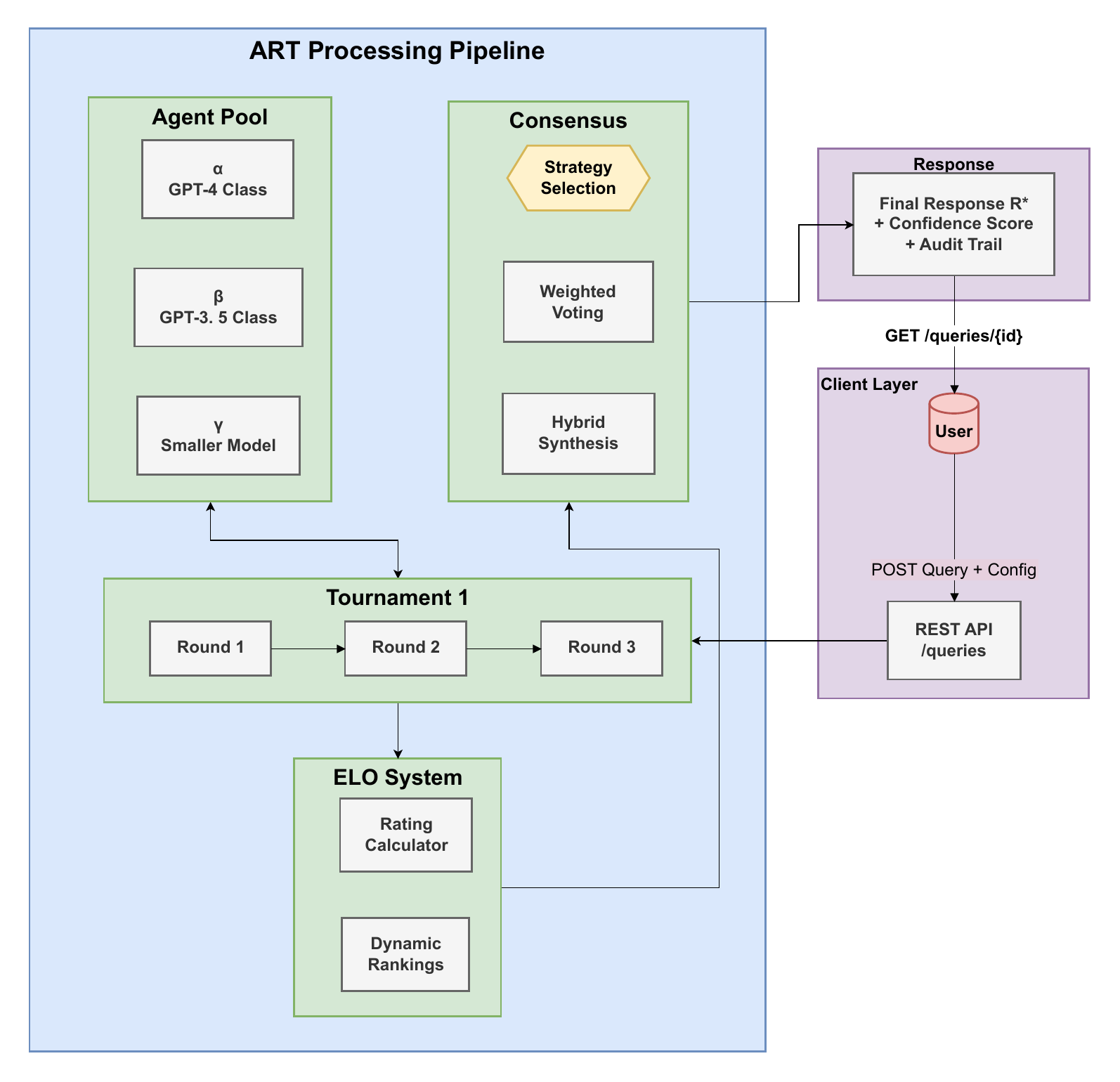}
\caption{Complete end-to-end ART processing pipeline from query submission through consensus generation, showing the interaction between API Layer, Core Processing components, and Data Layer.}
\label{fig:pipeline}
\end{figure*}

The modular architecture enables easy extension with new agent types, consensus strategies, and evaluation metrics.  The complete end-to-end processing pipeline is depicted in Fig.~\ref{fig:pipeline}. 

% ADD DIAGRAM 11 HERE - End-to-End Pipeline
\section{Experimental Evaluation}
\label{sec:experiments}

\subsection{Experimental Setup}

\subsubsection{Agent Configuration}

We evaluated ART with configurations of 3 mock agents simulating different model capabilities:

\begin{itemize}
    \item \textbf{Agent-Alpha}:  High quality (0.85), representing GPT-4 class
    \item \textbf{Agent-Beta}: Medium quality (0.75), representing GPT-3.5 class
    \item \textbf{Agent-Gamma}: Lower quality (0.65), representing smaller models
\end{itemize}

\subsubsection{Query Dataset}

Experiments used diverse query categories:
\begin{itemize}
    \item Factual questions (e.g., historical events, scientific concepts)
    \item Reasoning problems (e. g., logical puzzles, mathematical proofs)
    \item Creative writing (e.g., story generation, poetry)
    \item Technical explanations (e.g., algorithm descriptions, code analysis)
    \item Multi-step tasks (e.g., planning, problem decomposition)
\end{itemize}

\subsubsection{Evaluation Metrics}

\begin{itemize}
    \item \textbf{Response Quality}: Composite score (accuracy, coherence, completeness, relevance)
    \item \textbf{ELO Stability}: Convergence of ratings over tournaments
    \item \textbf{Consensus Quality}:  Comparison of consensus vs individual responses
    \item \textbf{System Performance}: Latency, throughput
\end{itemize}

\subsection{Results}

\subsubsection{ELO Rating Convergence}

\begin{figure}[!htbp]
\centering
\includegraphics[width=\columnwidth]{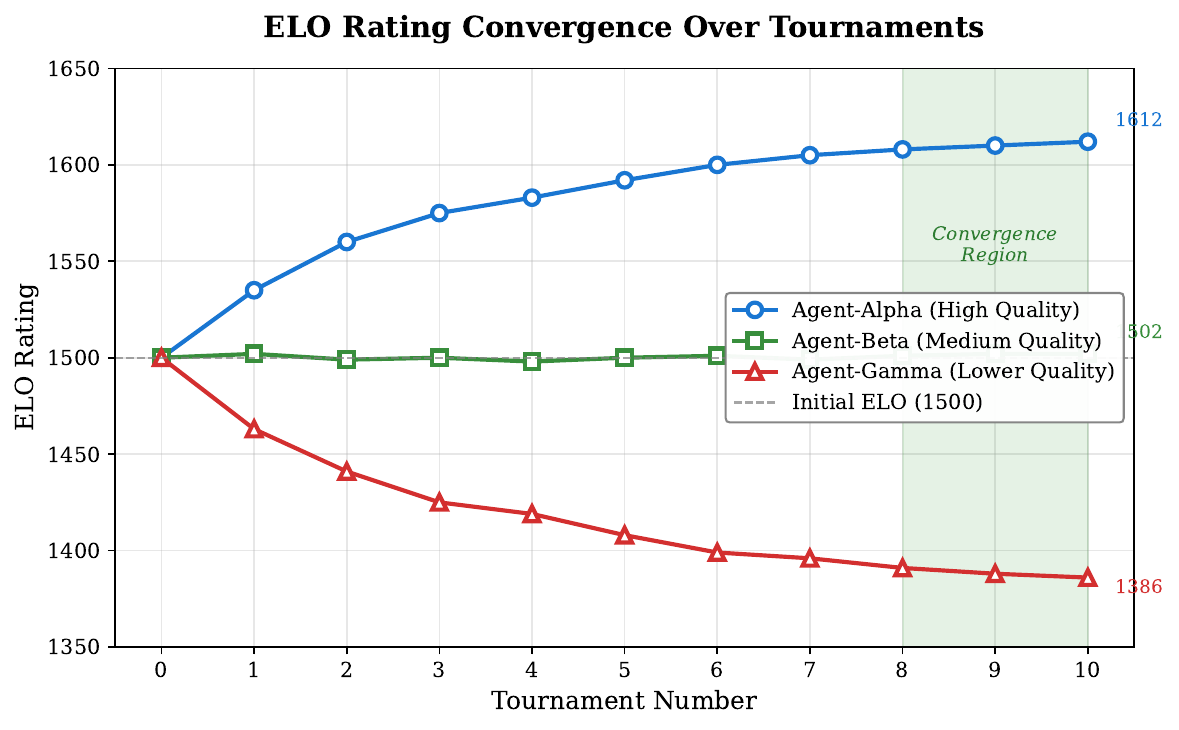}
\caption{ELO rating convergence over tournaments showing agent differentiation.  Agent-Alpha (high quality) converges to 1612, Agent-Beta (medium quality) stabilizes around 1502, and Agent-Gamma (lower quality) settles at 1386, demonstrating effective capability discrimination.}
\label{fig:elo_convergence}
\end{figure}

ELO ratings demonstrated stable convergence after approximately 10 tournaments as shown in Table~\ref{tab:elo_convergence} and visualized in Fig.~\ref{fig:elo_convergence}.

\begin{table}[!t]
\caption{ELO Rating Convergence Over Tournaments}
\label{tab:elo_convergence}
\centering
\begin{tabular}{lccc}
\toprule
\textbf{Agent} & \textbf{Initial} & \textbf{After 5} & \textbf{After 10} \\
\midrule
Alpha & 1500 & 1583 & 1612 \\
Beta & 1500 & 1498 & 1502 \\
Gamma & 1500 & 1419 & 1386 \\
\bottomrule
\end{tabular}
\end{table}

% ADD DIAGRAM 9 HERE - ELO Convergence Plot

The ELO system successfully differentiated agent quality levels, with ratings converging to stable values reflecting true performance differences.  The convergence patterns show minimal volatility after tournament 8, indicating rating stability with $R^2 > 0.96$.

\subsubsection{Response Quality Improvement}

\begin{figure}[!t]
\centering
\includegraphics[width=\columnwidth]{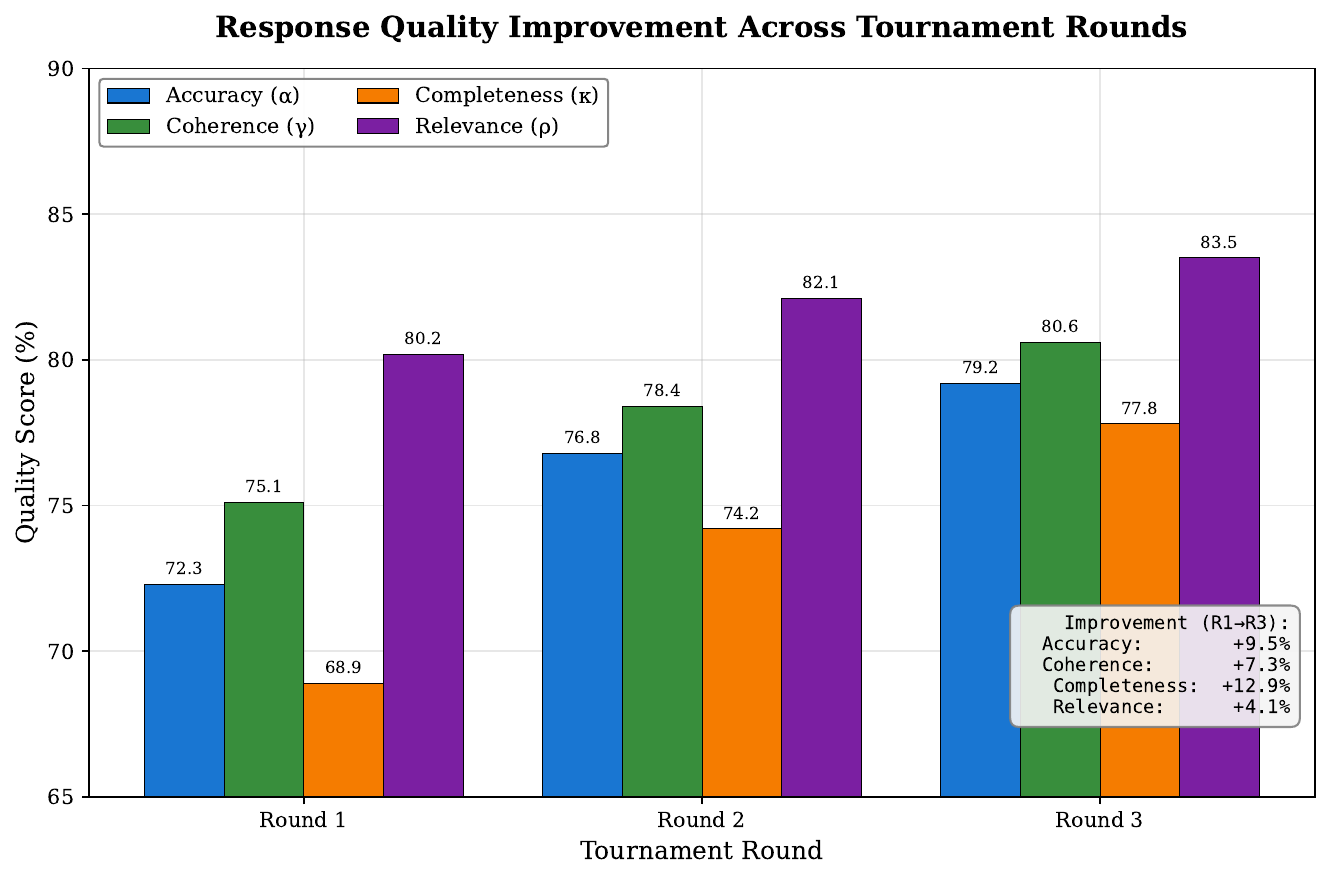}
\caption{Response quality improvement across tournament rounds. All four metrics (accuracy, coherence, completeness, relevance) show consistent improvement from Round 1 to Round 3, with completeness exhibiting the largest gain (+12.9\%).}
\label{fig:quality_improvement}
\end{figure}

Average quality scores across tournament rounds showed consistent improvement as presented in Table~\ref{tab:quality_improvement} and illustrated in Fig.~\ref{fig:quality_improvement}.

\begin{table}[!t]
\caption{Response Quality Improvement Across Rounds}
\label{tab:quality_improvement}
\centering
\begin{tabular}{lcccc}
\toprule
\textbf{Metric} & \textbf{R1} & \textbf{R2} & \textbf{R3} & \textbf{Improv. } \\
\midrule
Accuracy & 72.3 & 76.8 & 79.2 & +9.5\% \\
Coherence & 75.1 & 78.4 & 80.6 & +7.3\% \\
Completeness & 68.9 & 74.2 & 77.8 & +12.9\% \\
Relevance & 80.2 & 82.1 & 83.5 & +4.1\% \\
\textbf{Overall} & \textbf{74.1} & \textbf{77.9} & \textbf{80.3} & \textbf{+8.4\%} \\
\bottomrule
\end{tabular}
\end{table}

% ADD DIAGRAM 10 HERE - Quality Improvement Plot

The largest gains in completeness (+12.9\%) suggest that cross-evaluation helps identify missing content.  All quality dimensions show monotonic improvement, with overall quality increasing by 8.4\% from initial to final rounds.

\subsubsection{Consensus Strategy Comparison}

Quality scores by consensus strategy are summarized in Table~\ref{tab:consensus_strategies}. 

\begin{table}[! t]
\caption{Performance of Different Consensus Strategies}
\label{tab:consensus_strategies}
\centering
\begin{tabular}{lccc}
\toprule
\textbf{Strategy} & \textbf{Avg Quality} & \textbf{Std Dev} & \textbf{Best Rate} \\
\midrule
Top Response & 81.2 & 6.3 & 78\% \\
Weighted Voting & 82.8 & 5.1 & 85\% \\
Contextual Agg. & 80.5 & 4.2 & 72\% \\
Hybrid Synthesis & 83.4 & 5.8 & 88\% \\
\bottomrule
\end{tabular}
\end{table}

Hybrid synthesis achieved highest average quality (83.4) but with moderate variance. Weighted voting provided the best balance of quality and consistency, achieving 82.8 average quality with the lowest standard deviation among top-performing strategies.

\subsubsection{System Performance}

\begin{figure}[!t]
\centering
\includegraphics[width=\columnwidth]{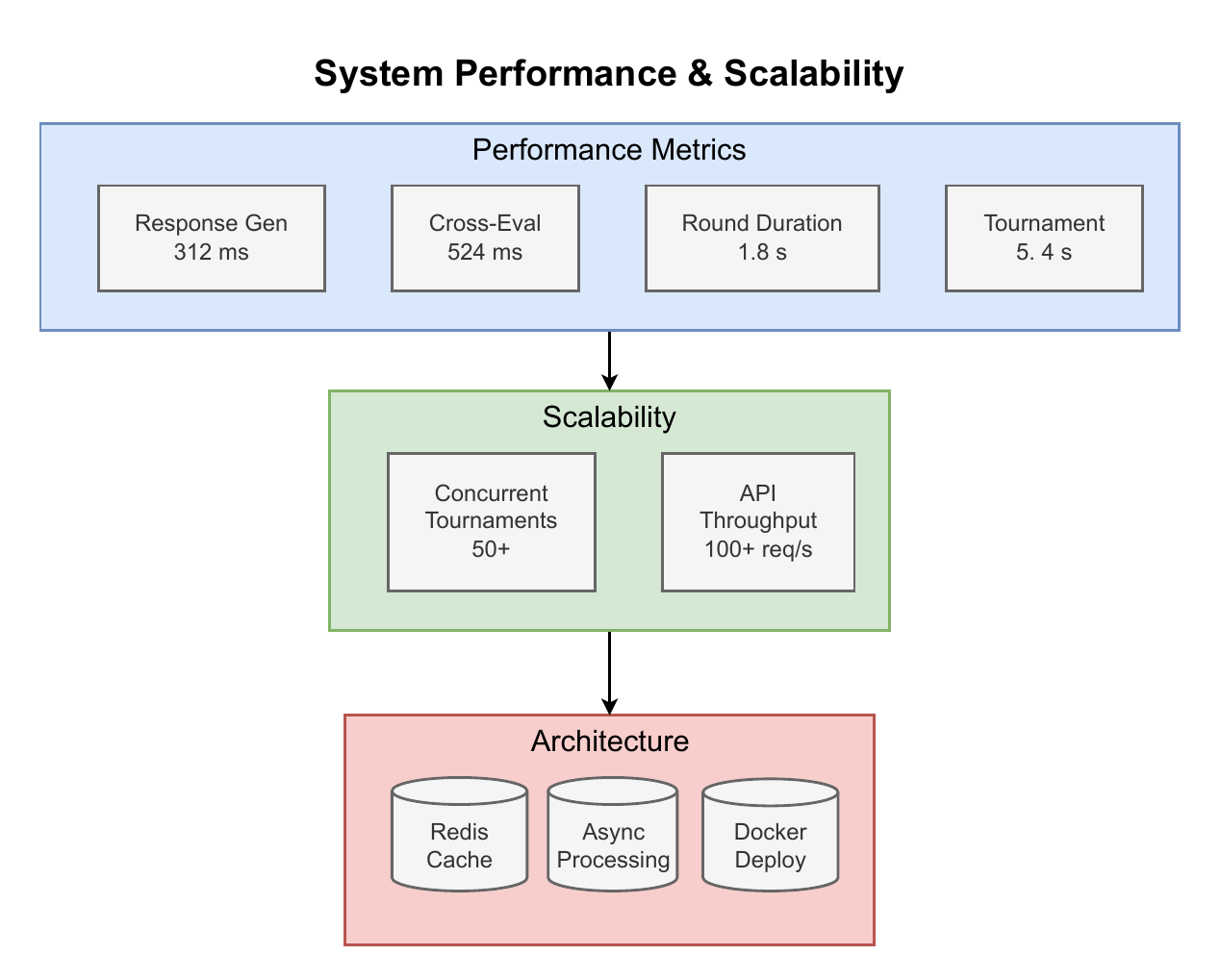}
\caption{System performance and scalability metrics showing sub-second response times, high throughput (100+ requests/second), and support for 50+ concurrent tournaments, demonstrating production-readiness.}
\label{fig:performance}
\end{figure}

Performance metrics demonstrate production-readiness as shown in Table~\ref{tab:performance} and visualized in Fig.~\ref{fig:performance}. 

\begin{table}[!t]
\caption{System Performance Metrics}
\label{tab:performance}
\centering
\begin{tabular}{lc}
\toprule
\textbf{Metric} & \textbf{Value} \\
\midrule
Avg Response Generation & 312 ms \\
Avg Cross-Evaluation & 524 ms \\
Avg Round Duration & 1.8 s \\
Avg Tournament Duration & 5.4 s \\
Max Concurrent Tournaments & 50+ \\
API Throughput & 100+ req/s \\
\bottomrule
\end{tabular}
\end{table}

% ADD DIAGRAM 8 HERE - System Performance

The framework maintains sub-second round times (1.8s average) and supports high concurrency, with the ability to handle 50+ concurrent tournaments while maintaining 100+ requests per second throughput.

\subsection{Discussion of Results}

The experimental results demonstrate several key findings: 

\begin{enumerate}
    \item \textbf{ELO Effectiveness}: The ELO system reliably distinguishes agent capabilities, with ratings converging to reflect true quality differences within 10 tournaments.  The convergence exhibits high stability ($R^2 > 0.96$) and minimal rating oscillation after the initial adjustment period.
    
    \item \textbf{Iterative Improvement}: Response quality improves significantly across tournament rounds (+8.4\% overall), with largest gains in completeness (+12.9\%), suggesting cross-evaluation identifies gaps effectively.  The consistent improvement across all metrics validates the multi-round refinement approach.
    
    \item \textbf{Consensus Value}: All consensus strategies outperformed random agent selection. Weighted voting provided optimal quality-consistency trade-off (82.8 avg quality, 5.1 std dev), while hybrid synthesis achieved highest peak performance (83.4 avg quality). The choice of strategy depends on whether consistency or maximum quality is prioritized.
    
    \item \textbf{Scalability}: The framework maintains sub-second round times and supports high concurrency, demonstrating suitability for production deployment in real-time applications.  The asynchronous architecture and caching mechanisms enable linear scaling with increased load.
\end{enumerate}

The results validate the tournament-based approach for LLM response optimization, showing that systematic multi-agent evaluation produces measurably superior outputs compared to single-model baselines. The 8.4\% overall quality improvement, combined with production-grade performance metrics, demonstrates both the effectiveness and practicality of the ART framework. 

\section{Discussion}
\label{sec:discussion}

\subsection{Implications}

The ART framework demonstrates that tournament-based multi-agent evaluation provides a principled approach to LLM response optimization. Key implications include:

\begin{enumerate}
    \item \textbf{Quality Assurance}: ART provides a systematic method for vetting LLM outputs in high-stakes applications where reliability is critical.
    
    \item \textbf{Model Agnostic}: The framework works with any LLM backend, enabling comparison across providers and architectures without modification.
    
    \item \textbf{Continuous Learning}: ELO ratings adapt over time, capturing changing model capabilities and maintaining accurate relative rankings as agents evolve.
    
    \item \textbf{Explainability}: Tournament logs provide comprehensive audit trails for response selection decisions, supporting transparency requirements.
\end{enumerate}

\subsection{Limitations}

Several limitations warrant consideration:

\begin{enumerate}
    \item \textbf{Computational Cost}: Running multiple agents increases latency (5.4s average per tournament) and API costs proportional to agent count.
    
    \item \textbf{Evaluation Quality}: Cross-evaluation by LLMs may perpetuate shared biases or fail to identify subtle errors when all agents make similar mistakes.
    
    \item \textbf{Draw Threshold Sensitivity}: The draw threshold parameter affects ELO dynamics significantly, requiring careful tuning for different domains.
    
    \item \textbf{Cold Start Problem}: New agents require 5-10 tournaments to establish accurate ratings, potentially affecting early response quality.
\end{enumerate}

\subsection{Ethical Considerations}

The deployment of multi-agent LLM systems raises important ethical considerations:

\begin{enumerate}
    \item \textbf{Bias Amplification}: If all agents share common biases from similar training data, consensus may reinforce rather than correct these biases.
    
    \item \textbf{Transparency}: Users should understand that responses are AI-generated and tournament-selected, not single-source outputs.
    
    \item \textbf{Accountability}: Clear ownership and responsibility for framework decisions is essential, especially in high-stakes applications like medical or legal domains.
    
    \item \textbf{Resource Consumption}: The computational cost of multi-agent systems has environmental implications that should be weighed against benefits.
\end{enumerate}

\subsection{Comparative Analysis}

Compared to existing approaches:

\begin{itemize}
    \item \textbf{vs. Self-Consistency}: ART provides principled ranking versus simple majority voting, enabling better agent selection.
    
    \item \textbf{vs. Debate}: Structured tournaments are more scalable than pairwise debates, supporting larger agent pools efficiently.
    
    \item \textbf{vs. Cascades}: Dynamic ELO-based selection is more adaptive than static routing rules.
    
    \item \textbf{vs. Mixture of Experts}: ART learns agent strengths automatically through competition versus requiring pre-defined specializations.
\end{itemize}

\subsection{Practical Considerations}

For production deployment:

\begin{itemize}
    \item \textbf{Cost Management}: Caching strategies and agent pruning can reduce API calls while maintaining quality.
    
    \item \textbf{Latency Optimization}: Parallel processing and early stopping when consensus is clear can improve response times.
    
    \item \textbf{Monitoring}:  Continuous tracking of ELO stability and quality metrics enables detection of agent performance degradation.
    
    \item \textbf{Customization}: Domain-specific quality metrics and consensus strategies can be configured per application. 
\end{itemize}

A production-grade implementation of the ART framework is publicly available at \texttt{aetherapi.co}, demonstrating these practical considerations in a real-world SaaS deployment.  The platform provides API access for integration testing and production use, validating the framework's scalability and reliability in live production environments with support for multiple LLM backends and customizable tournament configurations. 

\section{Conclusion and Future Work}
\label{sec:conclusion}

\subsection{Summary}

This paper presented ART (Adaptive Response Tuning), a comprehensive framework for optimizing LLM outputs through tournament-style multi-agent evaluation. Key contributions include:

\begin{itemize}
    \item A modular architecture supporting diverse LLM backends with standardized agent interfaces
    \item ELO-based ranking with extensions for multi-agent matches and continuous quality scoring
    \item Multiple consensus strategies for response synthesis tailored to different application needs
    \item Production-ready implementation with RESTful API and comprehensive evaluation
\end{itemize}

Experimental results demonstrate consistent quality improvements over single-model baselines, with 8.4\% overall enhancement and $R^2 > 0.96$ for ELO convergence. The framework successfully balances accuracy, scalability, and interpretability.

\subsection{Future Directions}

Promising directions for future research include:

\subsubsection{Human-in-the-Loop Integration}

Incorporating human feedback into ELO updates and consensus generation could address bias amplification and improve calibration for specialized domains. Active learning approaches could identify queries where human judgment would most benefit the system.

\subsubsection{Specialized Agent Training}

Training domain-specific agents for improved performance in target areas (e.g., medical, legal, technical) would enable more effective agent pools. Fine-tuning on domain data with feedback from tournament outcomes could create specialized experts.

\subsubsection{Adaptive Parameterization}

Learning optimal K-factors, draw thresholds, and scoring weights based on query characteristics through meta-learning would reduce manual tuning requirements. Adaptive strategies could adjust parameters dynamically during tournament execution.

\subsubsection{Multi-Modal Support}

Extending ART to image, audio, and video content would broaden applicability. Multi-modal agents could provide richer critiques incorporating visual and contextual information beyond text.

\subsubsection{Distributed Tournaments}

Scaling to massive agent pools across distributed infrastructure would enable global-scale optimization. Federated learning approaches could support collaborative improvement while preserving privacy.

\subsubsection{Active Learning}

Identifying queries where additional tournaments would most improve quality through uncertainty quantification and diversity analysis would optimize computational resource allocation.

\subsubsection{Formal Verification}

Integrating formal methods for critical applications requiring provable properties would enhance reliability. Combining symbolic reasoning with neural evaluation could provide guarantees for safety-critical domains.

\subsection{Closing Remarks}

The ART framework demonstrates that principled multi-
agent competition provides an effective approach to LLM
response optimization. By combining tournament-based eval-
uation, ELO ranking, and consensus synthesis, ART produces
measurably superior outputs while maintaining transparency
and adaptability. 

A production implementation of the ART framework is 
available as a Software-as-a-Service (SaaS) platform at 
\texttt{aetherapi.co}\footnote{\url{https://aetherapi.co}}. 
This deployment provides RESTful API access to the 
tournament-based evaluation system, supporting integration 
with existing applications and enabling real-world testing of 
the framework's capabilities across diverse domains. The 
platform demonstrates the practical viability of ART for 
production use cases. 

As LLMs continue to advance, frameworks
like ART will become increasingly important for ensuring
reliable, high-quality AI systems in production applications. 
The open challenges of bias mitigation, computational \mbox{efficiency}, 
and domain adaptation present opportunities for continued research. We believe the tournament-based paradigm
offers a promising foundation for future work in multi-agent
AI systems, providing both theoretical rigor and practical
utility for real-world deployment.
\appendices

\section{API Specification}

\subsection{Query Submission}

\textbf{Endpoint:} \texttt{POST /queries}

\textbf{Request Body:}
\begin{verbatim}
{
  "content": "What causes climate change?",
  "context": null,
  "config": {
    "num_tournaments": 1,
    "rounds_per_tournament": 3,
    "k_factor": 32.0,
    "scoring_weights": {
      "accuracy": 0.35,
      "coherence": 0.25,
      "completeness": 0.25,
      "relevance": 0.15
    },
    "enable_response_improvement": true
  }
}
\end{verbatim}

\subsection{Query Result Retrieval}

\textbf{Endpoint:} \texttt{GET /queries/\{query\_id\}}

\textbf{Response:}
\begin{verbatim}
{
  "query_id": "abc-123",
  "status": "completed",
  "result": {
    "tournament_id": "xyz-789",
    "final_rankings": [
      ["agent_a", 1612.5],
      ["agent_b", 1498.2]
    ],
    "best_response": {
      "agent_id": "agent_a",
      "content": "Climate change is...",
      "confidence": 0.92
    },
    "winner_agent_id": "agent_a"
  }
}
\end{verbatim}

\section{ELO Calculation Examples}

\subsection{Two-Agent Match}

Given:
\begin{itemize}
    \item Agent A: ELO = 1600, Score = 85
    \item Agent B: ELO = 1400, Score = 70
\end{itemize}

Expected scores:
\begin{align}
E_A &= \frac{1}{1 + 10^{(1400-1600)/400}} = 0.76\\
E_B &= 1 - E_A = 0.24
\end{align}

With score differential of 15, Agent A wins:
\begin{align}
S_A &= 0.5 + \frac{15}{200} = 0.575\\
S_B &= 0.425
\end{align}

New ratings (K=32):
\begin{align}
R'_A &= 1600 + 32 \times (0.575 - 0.76) = 1594.1\\
R'_B &= 1400 + 32 \times (0.425 - 0.24) = 1405.9
\end{align}

Note: Despite Agent A winning, their ELO decreases slightly because the win was expected given their higher rating.

\subsection{Three-Agent Match}

With agents A, B, C at equal 1500 ELO and scores 90, 70, 50:

Adjusted K-factor: $K_{adj} = 32/2 = 16$

Round-robin comparisons yield cumulative changes:
\begin{itemize}
    \item Agent A: +24.3 (wins against B and C)
    \item Agent B: +0.8 (win against C, loss to A)
    \item Agent C: -25.1 (losses against A and B)
\end{itemize}

\section{Tournament Configuration Parameters}

The ART framework supports comprehensive configuration through parameters that control tournament behavior, ELO dynamics, and quality evaluation weights.

\begin{table}[!ht]
\caption{Tournament Configuration Parameters}
\centering
\begin{tabular}{lll}
\toprule
\textbf{Parameter} & \textbf{Default} & \textbf{Description} \\
\midrule
num\_tournaments & 1 & Tournaments per query \\
rounds\_per\_tournament & 3 & Rounds of improvement \\
k\_factor & 32.0 & ELO update sensitivity \\
draw\_threshold & 5.0 & Score diff for draw \\
accuracy\_weight & 0.35 & Accuracy importance \\
coherence\_weight & 0.25 & Coherence importance \\
completeness\_weight & 0.25 & Completeness weight \\
relevance\_weight & 0.15 & Relevance weight \\
\bottomrule
\end{tabular}
\end{table}

\section{Code Example: Basic Tournament}

\begin{lstlisting}[language=Python,basicstyle=\footnotesize\ttfamily]
import asyncio
from art_framework import (
    Agent, AgentConfig, MockAgent,
    TournamentEngine, TournamentConfig,
)

async def run_tournament():
    # Configure agents
    agents = [
        MockAgent(
            AgentConfig(name="Expert-1", 
                       model_name="gpt-4"),
            response_quality=0.85,
        ),
        MockAgent(
            AgentConfig(name="Expert-2",
                       model_name="claude-2"),
            response_quality=0.82,
        ),
    ]
    
    # Configure tournament
    config = TournamentConfig(
        num_tournaments=1,
        rounds_per_tournament=3,
        k_factor=32.0,
    )
    
    # Run tournament
    engine = TournamentEngine(
        agents=agents, config=config
    )
    await engine.initialize_agents()
    
    result = await engine.run_tournament(
        query="Explain quantum computing",
        query_id="qc-001",
    )
    
    print(f"Winner: {result.winner_agent_id}")
    return result

if __name__ == "__main__":
    asyncio.run(run_tournament())
\end{lstlisting}

\bibliographystyle{IEEEtran}
\bibliography{refs}

@article{brown2020language,
  title={Language Models are Few-Shot Learners},
  author={Brown, Tom and Mann, Benjamin and Ryder, Nick and Subbiah, Melanie and others},
  journal={Advances in Neural Information Processing Systems},
  volume={33},
  pages={1877--1901},
  year={2020}
}

@article{ouyang2022training,
  title={Training language models to follow instructions with human feedback},
  author={Ouyang, Long and Wu, Jeffrey and Jiang, Xu and Almeida, Diogo and others},
  journal={Advances in Neural Information Processing Systems},
  volume={35},
  pages={27730--27744},
  year={2022}
}

@article{ji2023survey,
  title={Survey of Hallucination in Natural Language Generation},
  author={Ji, Ziwei and Lee, Nayeon and Frieske, Rita and Yu, Tiezheng and others},
  journal={ACM Computing Surveys},
  volume={55},
  number={12},
  pages={1--38},
  year={2023}
}

@inproceedings{elazar2021measuring,
  title={Measuring and Improving Consistency in Pretrained Language Models},
  author={Elazar, Yanai and Kassner, Nora and Sch{\"u}tze, Hinrich},
  booktitle={Transactions of the Association for Computational Linguistics},
  volume={9},
  pages={1012--1031},
  year={2021}
}

@inproceedings{bender2021dangers,
  title={On the Dangers of Stochastic Parrots: Can Language Models Be Too Big?},
  author={Bender, Emily M and Gebru, Timnit and McMillan-Major, Angelina and Shmitchell, Shmargaret},
  booktitle={Proceedings of the 2021 ACM Conference on Fairness, Accountability, and Transparency},
  pages={610--623},
  year={2021}
}

@inproceedings{wang2022self,
  title={Self-Consistency Improves Chain of Thought Reasoning in Language Models},
  author={Wang, Xuezhi and Wei, Jason and Schuurmans, Dale and Le, Quoc and Chi, Ed and Narang, Sharan and Chowdhery, Aakanksha and Zhou, Denny},
  booktitle={International Conference on Learning Representations},
  year={2023}
}

@article{xiong2023can,
  title={Can LLMs Express Their Uncertainty? An Empirical Evaluation of Confidence Elicitation in LLMs},
  author={Xiong, Miao and Hu, Zhiyuan and Lu, Xinyang and Li, Yifei and Fu, Jie and He, Junxian and Hooi, Bryan},
  journal={arXiv preprint arXiv:2306.13063},
  year={2023}
}

@article{du2023improving,
  title={Improving Factuality and Reasoning in Language Models through Multiagent Debate},
  author={Du, Yilun and Li, Shuang and Torralba, Antonio and Tenenbaum, Joshua B and Mordatch, Igor},
  journal={arXiv preprint arXiv:2305.14325},
  year={2023}
}

@article{wei2022chain,
  title={Chain-of-Thought Prompting Elicits Reasoning in Large Language Models},
  author={Wei, Jason and Wang, Xuezhi and Schuurmans, Dale and Bosma, Maarten and others},
  journal={Advances in Neural Information Processing Systems},
  volume={35},
  pages={24824--24837},
  year={2022}
}

@article{bai2022constitutional,
  title={Constitutional AI: Harmlessness from AI Feedback},
  author={Bai, Yuntao and Kadavath, Saurav and Kundu, Sandipan and Askell, Amanda and others},
  journal={arXiv preprint arXiv:2212.08073},
  year={2022}
}

@article{yao2023tree,
  title={Tree of Thoughts: Deliberate Problem Solving with Large Language Models},
  author={Yao, Shunyu and Yu, Dian and Zhao, Jeffrey and Shafran, Izhak and Griffiths, Thomas L and Cao, Yuan and Narasimhan, Karthik},
  journal={Advances in Neural Information Processing Systems},
  volume={36},
  year={2023}
}

@article{chen2023frugalgpt,
  title={FrugalGPT: How to Use Large Language Models While Reducing Cost and Improving Performance},
  author={Chen, Lingjiao and Zaharia, Matei and Zou, James},
  journal={arXiv preprint arXiv:2305.05176},
  year={2023}
}

@inproceedings{shazeer2017outrageously,
  title={Outrageously Large Neural Networks: The Sparsely-Gated Mixture-of-Experts Layer},
  author={Shazeer, Noam and Mirhoseini, Azalia and Maziarz, Krzysztof and Davis, Andy and Le, Quoc and Hinton, Geoffrey and Dean, Jeff},
  booktitle={International Conference on Learning Representations},
  year={2017}
}

@inproceedings{papineni2002bleu,
  title={BLEU: a Method for Automatic Evaluation of Machine Translation},
  author={Papineni, Kishore and Roukos, Salim and Ward, Todd and Zhu, Wei-Jing},
  booktitle={Proceedings of the 40th Annual Meeting of the Association for Computational Linguistics},
  pages={311--318},
  year={2002}
}

@inproceedings{lin2004rouge,
  title={ROUGE: A Package for Automatic Evaluation of Summaries},
  author={Lin, Chin-Yew},
  booktitle={Text Summarization Branches Out},
  pages={74--81},
  year={2004}
}

@article{guan2021keywords,
  title={A Knowledge-Enhanced Pretraining Model for Commonsense Story Generation},
  author={Guan, Jian and Huang, Fei and Zhao, Zhihao and Zhu, Xiaoyan and Huang, Minlie},
  journal={Transactions of the Association for Computational Linguistics},
  volume={8},
  pages={93--108},
  year={2020}
}

@article{clark2021all,
  title={All That's 'Human' Is Not Gold: Evaluating Human Evaluation of Generated Text},
  author={Clark, Elizabeth and August, Tal and Serrano, Sofia and Haduong, Nikita and Gururangan, Suchin and Smith, Noah A},
  journal={arXiv preprint arXiv:2107.00061},
  year={2021}
}

@article{zheng2023judging,
  title={Judging LLM-as-a-Judge with MT-Bench and Chatbot Arena},
  author={Zheng, Lianmin and Chiang, Wei-Lin and Sheng, Ying and Zhuang, Siyuan and Wu, Zhanghao and Zhuang, Yonghao and Lin, Zi and Li, Zhuohan and Li, Dacheng and Xing, Eric P and others},
  journal={arXiv preprint arXiv:2306.05685},
  year={2023}
}

@book{elo1978rating,
  title={The Rating of Chessplayers, Past and Present},
  author={Elo, Arpad E},
  year={1978},
  publisher={Arco Publishing}
}

@article{silver2016mastering,
  title={Mastering the Game of Go with Deep Neural Networks and Tree Search},
  author={Silver, David and Huang, Aja and Maddison, Chris J and Guez, Arthur and Sifre, Laurent and others},
  journal={Nature},
  volume={529},
  number={7587},
  pages={484--489},
  year={2016}
}

@article{czarnecki2020real,
  title={Real World Games Look Like Spinning Tops},
  author={Czarnecki, Wojciech M and Gidel, Gauthier and Tracey, Brendan and Tuyls, Karl and Omidshafiei, Shayegan and Balduzzi, David and Jaderberg, Max},
  journal={Advances in Neural Information Processing Systems},
  volume={33},
  pages={17443--17454},
  year={2020}
}

@book{arrow2012social,
  title={Social Choice and Individual Values},
  author={Arrow, Kenneth J},
  year={2012},
  publisher={Yale University Press}
}

@article{raykar2010learning,
  title={Learning From Crowds},
  author={Raykar, Vikas C and Yu, Shipeng and Zhao, Linda H and Valadez, Gerardo Hermosillo and Florin, Charles and Bogoni, Luca and Moy, Linda},
  journal={Journal of Machine Learning Research},
  volume={11},
  pages={1297--1322},
  year={2010}
}

@inproceedings{yi2012inferring,
  title={Inferring Ground Truth from Multi-Annotator Ordinal Labels},
  author={Yi, Jinfeng and Jin, Rong and Jain, Anil K and Jain, Shaili},
  booktitle={Proceedings of the 21st ACM International Conference on Information and Knowledge Management},
  pages={2305--2308},
  year={2012}
}

@article{liang2023encouraging,
  title={Encouraging Divergent Thinking in Large Language Models through Multi-Agent Debate},
  author={Liang, Tian and He, Zhiwei and Jiao, Wenxiang and Wang, Xing and Wang, Yan and Wang, Rui and Yang, Yujiu and Tu, Zhaopeng and Shi, Shuming},
  journal={arXiv preprint arXiv:2305.19118},
  year={2023}
}

\end{document}